%% file: 0_ntg-cvpr19.tex
\documentclass[10pt,twocolumn,letterpaper]{article}
\usepackage[pagebackref=true,breaklinks=true,letterpaper=true,colorlinks,bookmarks=false]{hyperref}
\usepackage{cvpr}
\usepackage{times}
\usepackage{epsfig}
\usepackage{graphicx}
\usepackage{amsmath}
\usepackage{amssymb}

\usepackage{color}
\usepackage{xspace}

\newcommand{\NE}{NE}
\newcommand{\figref}[1]{{Figure \ref{fig:#1}}}
\newcommand{\secref}[1]{{Section \ref{sec:#1}}}

\newcommand{\exec}{execution engine\xspace}

\cvprfinalcopy % *** Uncomment this line for the final submission

\def\cvprPaperID{864} % *** Enter the CVPR Paper ID here

\ifcvprfinal\fi
\begin{document}

%%%%%%%%%%%%%%%%%%%%%%%%%%%%%%%%%%%%%%%%%%%%%%%%%%%%%%%%%%%%%%%%%%%%%%%%%%%%%%
% space tweaks 
%%%%%%%%%%%%%%%%%%%%%%%%%%%%%%%%%%%%%%%%%%%%%%%%%%%%%%%%%%%%%%%%%%%%%%%%%%%%%%
% \setlength{\abovecaptionskip}{0.5mm}
% \setlength{\belowcaptionskip}{1.0mm} 
% \setlength{\textfloatsep}{1.5mm}
% \setlength{\dbltextfloatsep}{1.5mm}
%\renewcommand{\baselinestretch}{.9945} %not needed since references are not restricted

%%%%%%%%% TITLE
\title{Neural Task Graphs: Generalizing to Unseen Tasks \\from a Single Video Demonstration}

\author{De-An Huang*, Suraj Nair*, Danfei Xu*, Yuke Zhu, Animesh Garg,\\ Li Fei-Fei, Silvio Savarese, Juan Carlos Niebles\\
Computer Science Department,
	Stanford University
}

\maketitle
%\thispagestyle{empty}

%%%%%%%%% ABSTRACT
\begin{abstract}

Our goal is to generate a policy to complete an unseen task given just a single video demonstration of the task in a given domain. We hypothesize that to successfully generalize to unseen complex tasks from a single video demonstration, it is necessary to explicitly incorporate the compositional structure of the tasks into the model. To this end, we propose Neural Task Graph (NTG) Networks, which use conjugate task graph as the intermediate representation to modularize both the video demonstration and the derived policy. We empirically show NTG achieves inter-task generalization on two complex tasks: Block Stacking in BulletPhysics and Object Collection in AI2-THOR. NTG improves data efficiency with visual input as well as achieve strong generalization without the need for dense hierarchical supervision. We further show that similar performance trends hold when applied to real-world data. We show that NTG can effectively predict task structure on the JIGSAWS surgical dataset and generalize to unseen tasks.
\renewcommand*{\thefootnote}{}
\footnote{* indicates equal contribution}
\renewcommand*{\thefootnote}{\arabic{footnote}}
\setcounter{footnote}{0}
\vspace{-2mm}
\end{abstract}

\input{1_intro-v2.tex}

\input{2_related.tex}

\input{3_method.tex}

\input{4_experiment.tex}

\input{5_conclusion.tex}

\vspace{2mm}
\noindent\textbf{Acknowledgements.} Toyota Research Institute (``TRI'')  provided funds to assist the authors with their research but this article solely reflects the opinions and conclusions of its authors and not TRI or any other Toyota entity.

{\small
\bibliographystyle{ieee}
\bibliography{ntg-bib}
}

\end{document}

%% file: 1_intro-v2.tex
\section{Introduction}

% One-shot imitation learning
Learning sequential decisions and adapting to new task objectives at test time is a long-standing challenge in AI~\cite{brooks1986robust,fikes1972learning}. In rich real domains, an autonomous agent has to acquire new skills with minimal supervision. 
Recent works have tackled the problem of one-shot imitation learning~\cite{duan2017one,finn2017one,wu2010towards,xu2017neural} that learns from a single demonstration. In this work, we push a step further to address one-shot \emph{visual} imitation learning that operates directly on \emph{videos}.
We first train a model on a set of seen in-domain tasks. The model can then be applied on a single video demonstration to obtain an execution policy of the new unseen task.

Learning directly from video is crucial for advancing the existing imitation learning approaches to real-world scenarios as it is infeasible to annotate states, such as object trajectories, in each video. We focus on \emph{long-horizon tasks}, as real-world tasks such as cooking or assembly are inherently long-horizon and hierarchical.
Recent works have attempted learning from pixel space~\cite{finn2017one,liu2017imitation,sermanet2017time,yu2018one}, but learning long-horizon tasks from video in a one-shot setting remains a challenge,
since both the visual learning and task complexity exacerbate the demand for better data efficiency.

\begin{figure}[t]
    \centering
    \includegraphics[width=\linewidth]{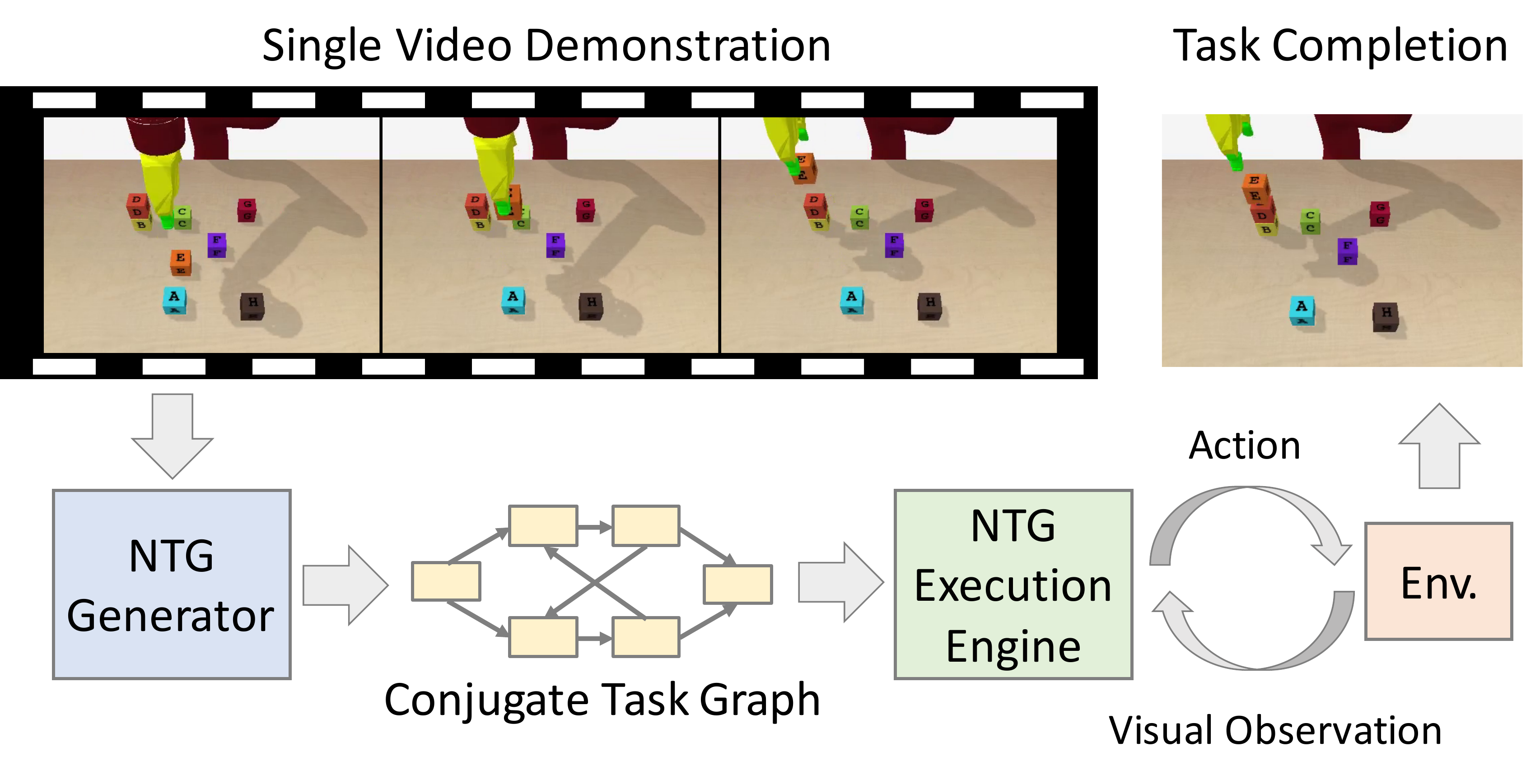}
    \caption{Our goal is to execute an unseen task from a single video demonstration. We propose Neural Task Graph Networks that leverage compositionality by using the task graph as the intermediate representation. This leads to strong inter-task generalization. 
    }
    \label{fig:fig1}
       \vspace{-2mm}
\end{figure}

Our solution explicitly models the compositionality in the task structure and policy, enabling us to scale one-shot visual imitation to complex tasks.
This is in contrast to previous works using unstructured task representations and policies~\cite{duan2017one,finn2017one}. The use of compositionality has led to better generalization in Visual Question Answering~\cite{hu2017learning,johnson2017inferring,kottur2018visual} and Policy Learning~\cite{andreas2016modular,devin2017learning,wang2018nervenet}. 
We propose Neural Task Graph (NTG) Networks, a novel framework that uses task graph as the intermediate representation to explicitly modularize both the visual demonstration and the derived policy.  NTG consists of a generator and an \exec, where the generator builds a \emph{task graph} from the task demo video to capture the structure of the task,
and the \exec interacts with the environment to perform the task conditioned on the inferred task graph. 
\figref{fig1} shows an overview of NTG Networks.

The main technical challenge in using graphical task representations is that the unseen demos can easily introduce states that are never observed during training.
For example, the goal state of an unseen block stacking task~\cite{duan2017one,xu2017neural} is a block configuration that never appears during training.
This challenge is amplified by our goal of learning from visual observation without strong supervision,
which obscures the state structure and prevents direct state space decomposition, as done in prior work~\cite{duan2017one}. 
Our key observation is that, {while there can be countless possible states, the number of possible actions in a certain domain is often limited.}
We leverage this conjugate relationship between states and actions, and propose to learn NTG on the Conjugate Task Graph (CTG)~\cite{hayes2016autonomously}, where the nodes are actions, and the states are captured by the edges. This allows us to modularize the policy and address the challenge of an unknown number of novel states. This is critical when operating in visual space, where states are high dimensional images and modeling a graph over a combinatorial state space is infeasible. Additionally, the CTG intermediate representation can yield alternate action sequences to complete the task, a property that is vital for generalization to unseen scenarios in a world with stochastic dynamics. This sets NTG apart from previous works that directly output the policy over options~\cite{xu2017neural} or actions~\cite{duan2017one} from a single demonstration.

We evaluate NTG Networks on one-shot visual imitation learning in two domains: Block Stacking in a robot simulator~\cite{BulletPhysics} and Object Collection in AI2-THOR~\cite{ai2thor}. Both domains involve multi-step planning for interaction and are inherently compositional. 
We show that NTG significantly improves the data efficiency on these complex tasks for direct imitation from video by explicitly incorporating compositionality. 
We also show that with the data-driven task structure, NTG outperforms methods that learn unstructured task representation~\cite{duan2017one} and methods that use strong hierarchically structured supervision~\cite{xu2017neural}, albeit without requiring detailed supervision. 
Further, we evaluate NTG on real-world videos. We show that NTG can effectively predict task graph structure on the JIGSAWS \cite{Gao2014JHUISIGA} surgical dataset and generalize to unseen human demonstrations.

In summary, the main contributions of our work are: 
(1) Introducing compositionality to both the task and policy representation to enable one-shot visual imitation learning on long-horizon tasks; (2) Proposing Neural Task Graph (NTG) Networks, a novel framework that uses task graph to capture the structure and the goal of a task;
(3) Addressing the challenge of novel visual state decomposition using a Conjugate Task Graph (CTG) formulation.

%% file: 2_related.tex
\section{Related Work}
% \todo{Rewrite everything.}

\noindent \textbf{Imitation Learning.} 
Traditional imitation learning work uses physical guidance~\cite{akgun2012keyframe,niekum2012learning} or teleoperation~\cite{whitney17a,zhang2017deep}  as demonstration.
While, third-person imitation learning uses date from other agents or viewpoints~\cite{liu2017imitation,sermanet2017time}. 
Recent methods for one-shot imitation learning~\cite{duan2017one,finn2017one,goo2018one,wu2010towards,xu2017neural,yu2018one} can translate a single demonstration to an executable policy. The most similar to ours is NTP~\cite{xu2017neural} that also learns long-horizon tasks. However, NTP (1) uses strong hierarchical frame label supervision and (2) suffers from a noticeable drop in performance with visual state. Our method reduces the need for this strong supervision, requiring only the demonstration action sequence during training, while achieving a performance boost of over 25\% in success rates.

\begin{figure}[t]
    \centering
    \includegraphics[width=0.85\linewidth]{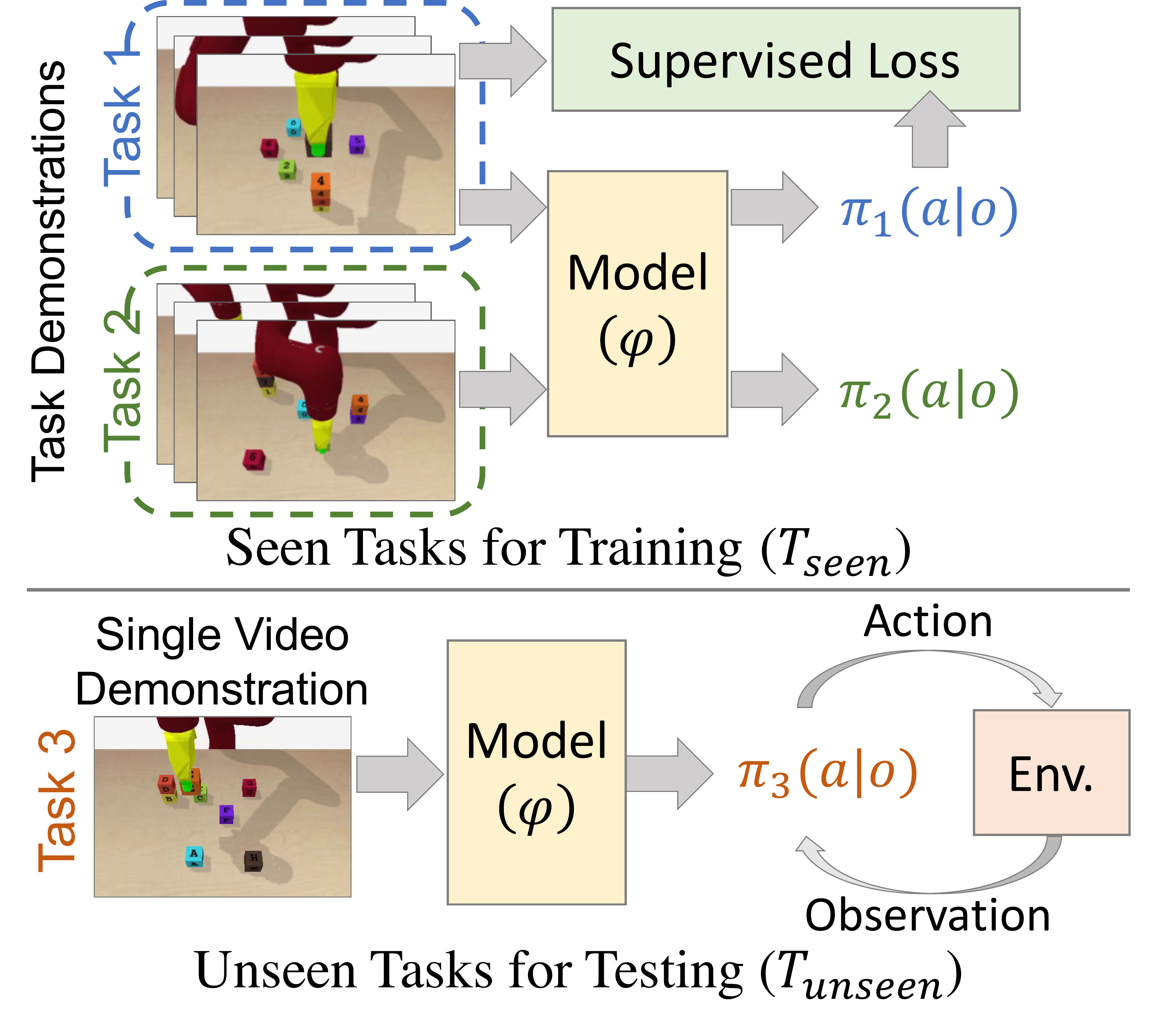}
    \caption{Overview of the setting of one-shot visual imitation learning. The seen tasks (Task 1 and 2) are used to train the model $\phi$ to instantiate the policy $\pi_i$ from the demonstration. During testing, $\phi$ is applied to a single video demonstration from the unseen Task 3 to generate the policy $\pi_3$ to interact with the environment. }
    \label{fig:one_shot}
\end{figure}

\begin{figure*}[t]
    \centering
    \includegraphics[width=0.90\linewidth]{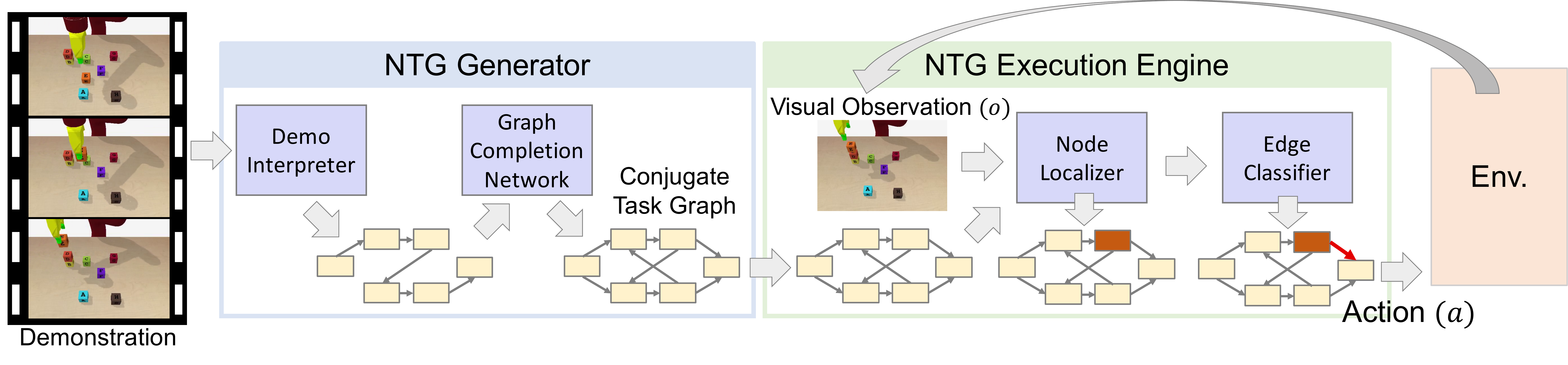}
    \caption{Overview of our Neural Task Graph (NTG) networks. The NTG networks consist of a generator that produces the conjugate task graph as the intermediate representation, and an \exec that executes the graph by localizing node and deciding the edge transition in the task graph based on the current visual observation.}
    \label{fig:sys_fig}
\end{figure*}

\vspace{1mm}
\noindent \textbf{Task Planning and Representations.} 
Conventionally task planning focuses on high-level plans and low-level state spaces~\cite{fikes1971strips,srivastava2014combined}. 
Recent works integrate perception via deep  learning~\cite{gupta2017cognitive,pinto2016curious,zhu2017visual}. 
HTN compounds low-level sub-tasks into higher-level abstraction to reduce the planning complexity~\cite{nau1999shop,sacerdoti1975structure}. 
Other representations include: integrating task and motion planning~\cite{kaelbling2011hierarchical} and  behavior-based systems~\cite{nicolescu2002hierarchical}. In vision, And-Or Graphs capture the hierarchical structures and have been used to parse video demonstrations~\cite{liu2016jointly}. 
Unlike previous methods, our task graph representation is data-driven and domain-agnostic: we generate nodes and edges directly from task demonstrations.

\vspace{1mm}
\noindent \textbf{Structural Video Understanding.} 
Generating task graphs from demonstrations is related to video understanding. 
Annotation in videos is hard to obtain. One solution is to use the language as supervision. This includes instructional video~\cite{alayrac16unsupervised,huang2018finding,sener2015unsupervised}, movie script~\cite{tapaswi2015book2movie,zhu2015aligning}, and caption annotation~\cite{gupta2009understanding,krishna2017dense}. We focus on how the structure is helpful for task learning, and assume the annotation for the seen tasks.

\vspace{1mm}
\noindent \textbf{Compositional Models in Vision and Robotics.} 
Recent works have utilized compositionality to improve models' generalization,
including visual question answering~\cite{andreas2015deep,hu2017learning,johnson2017inferring} and policy learning~\cite{andreas2016modular}. 
We show the same principle can significantly improve data efficiency in imitation learning to enable visual learning of complex tasks.

%% file: 3_method.tex
\section{Problem Formulation}  % TODO
\label{sec:problem}

Our goal is to learn to execute a previously unseen task from a single video demonstration. We refer to this as one-shot \emph{visual} imitation to emphasize that the model directly learns from visual inputs.
Let $\mathbb{T}$ be the set of all tasks in the domain of interest, $\mathbb{A}$ be the set of high-level actions, and $\mathbb{O}$ be the space of visual observation. 
A video demonstration $d$ for a task $\tau$ is defined as a video, $d^{\tau} = [o_1, \dots, o_T]$, that complete the task.
As shown in \figref{one_shot}, in one-shot imitation learning, $\mathbb{T}$ is split into two sets: $\mathbb{T}_{seen}$ with a large amount of demonstrations and supervision for training, and $\mathbb{T}_{unseen}$ with only task demonstrations for evaluation.
The goal is to learn a model $\phi(\cdot)$ from $\mathbb{T}_{seen}$ that can instantiate a policy $\pi_d(a|o)$ from the demonstration $d$ to perform the tasks in $\mathbb{T}_{unseen}$ based on the visual observation $o$.

The learning problem is formulated as learning a model $\phi(\cdot)$ that maps demonstration $d$ to the policy $\phi(d) = \pi_d(a|o)$. $\mathbb{T}_{seen}$ is used to train this model with demonstrations and potentially extra supervision. 
At test time, given a demonstration $d$ from an unseen task, the hope is that $\phi(\cdot)$ trained on $\mathbb{T}_{seen}$ can generalize to novel task instances in $\mathbb{T}_{unseen}$ and produce a  policy that can complete the novel task illustrated by the visual demonstration.

\section{Neural Task Graph Networks}

We have formulated one-shot visual imitation as learning the model $\phi(\cdot)$ that instantiates a policy from a single video demonstration. 
As shown in \figref{fig1}, our key contribution is explicitly incorporating \emph{compositionality} to $\phi(\cdot)$ to improve data efficiency of generalization.
We decompose $\phi(\cdot)$ into two components: a \emph{graph generator} $\phi_{gen}(\cdot)$ for generating the task graph $G$ from the demonstration ($G = \phi_{gen}(d)$), and a \emph{graph \exec} $\phi_{exe}(\cdot)$ that executes the task graph and acts as the policy ($\pi_d = \phi_{exe}(G)$). The structure of the task graph $G$ provides compositionality for both the demonstration and the policy. This leads to stronger data efficiency of generalizing to unseen tasks. An overview of our model is shown in \figref{sys_fig}.

\subsection{Neural Task Graph Generator}
\label{sec:generator}

The NTG Generator generates a task  graph capturing the structure of an unseen task from a single video demonstration. This is challenging since the video demonstration of an unseen task introduces novel visual states that are not observed in the seen tasks. This challenge is amplified by our goal of learning from  \emph{visual} observation, which prevents direct state space decomposition. In this case, generating the traditional task graph is ill-posed due to the exploding number of nodes. We address this by leveraging the conjugate relationship between state and action and work with the conjugate task graph~\cite{hayes2016autonomously}, where the nodes are the actions, and the edges implicitly depend on the current state. In the experiments, we show that this scheme significantly simplifies the (conjugate) task graph generation problem.

\vspace{1mm}
\noindent\textbf{Conjugate Task Graph (CTG).} 
% We follow the definitions by Hayes and Scassellati~\cite{hayes2016autonomously}: 
A \emph{task graph} $\bar{G} = \{\bar{V},\bar{E}\}$ contains nodes $\bar{V}$ as the states and $\bar{E}$ the directed edges for the transitions or actions between them. A successful execution of the task is equivalent to a path in the graph that reaches the goal node. The task graph captures the structure of the task, and the effect of each action. %how each action would affect the environment state.
However, generating this graph for an unseen task is extremely challenging, as each unseen state would be mapped to a new node. This is especially the case in visual tasks, where the state space is high dimensional.  We thus work with the \emph{conjugate task graph}(CTG)~\cite{hayes2016autonomously}, ${G} = \{{V},{E}\}$, where the actions are now the nodes $V$, 
and the states become edges $E$, which implicitly encode the preconditions of the actions.
This allows us to bypass explicit state modeling, while still being able to perform tasks by traversing the conjugate task graph.
% Depending on the context, % in the sequel,  we may just refer to the conjugate task graph as the task graph.

We assume that all actions are observed during training from the seen tasks, which is reasonable for tasks in the same domain. This gives all the nodes in CTG,
and the goal is to infer the correct edges. This can be viewed as understanding the preconditions for each action.
We propose two steps for generating the edges: 
(i) \emph{Demo Interpretation}: First we obtain a valid path traversing the conjugate task graph by observing the action order in the demonstration;
(ii) \emph{Graph Completion}: The second step is to add the edges that are not observed in the demonstration. There might be actions whose order can be permutated without affecting the final outcome. As we only have a single demonstration, this interchangeability is not captured in the previous step. We learn a Graph Completion Network, which adds more edges that are proper given the edges initialized by step (i).

\begin{figure}[t!]
    \centering
    \includegraphics[width=\linewidth]{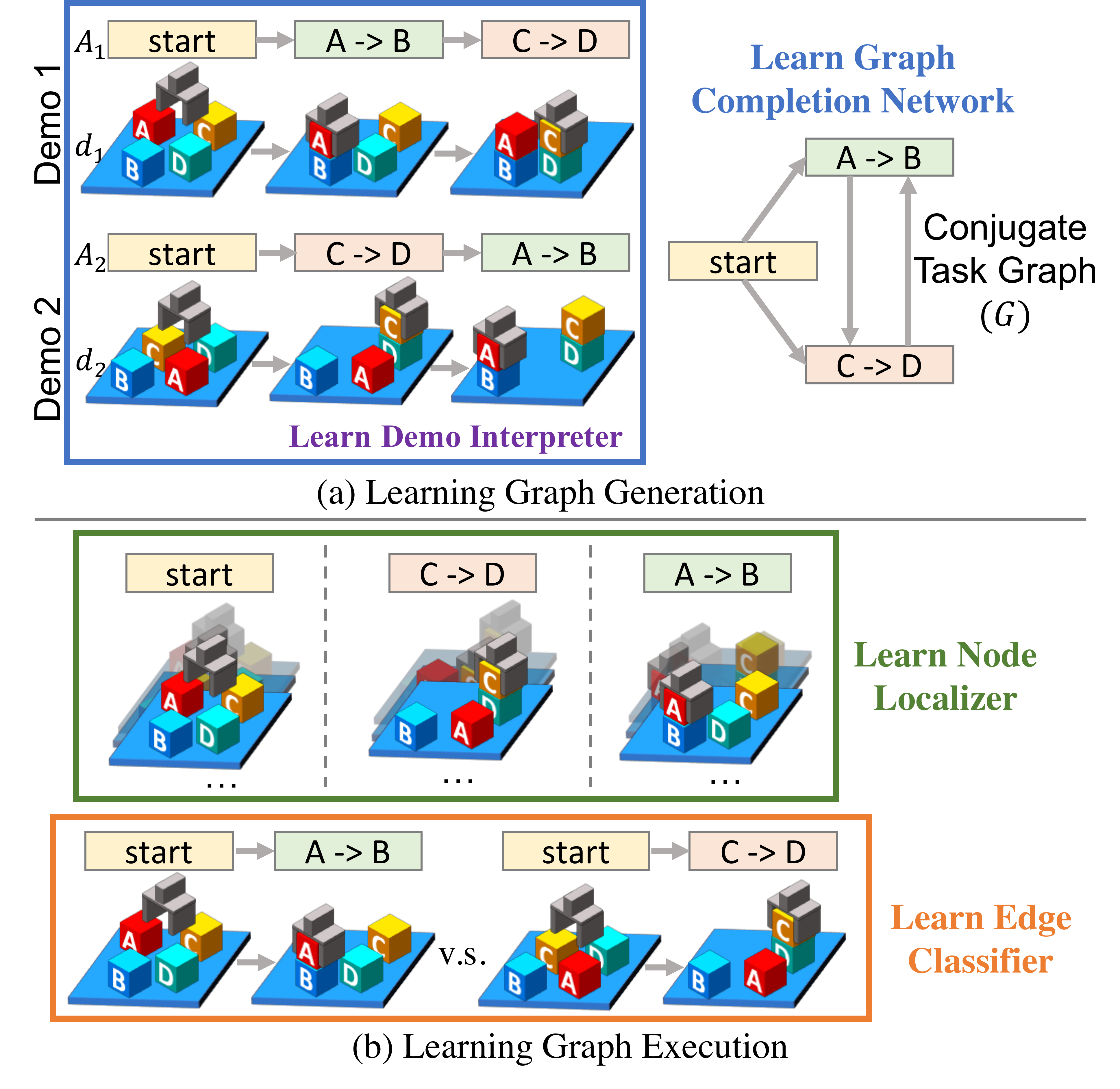}
    \caption{ 
    Illustration of our learning setting with a block stacking task.
    The video demonstrations $d_i$ in the seen tasks only require corresponding action sequence $A_i$. We aggregate data from all the demonstrations in the same task and use it as the supervision of each component of our model. This approach allows us to bypass the need for strong supervision as in previous works.
    }
    \label{fig:learning}
\end{figure}

\vspace{1mm}
\noindent\textbf{Demo Interpreter.} 
Given the demonstration $d = [o_1, \dots, o_T]$, our goal is to output $A = [a_1, \dots, a_K]$, the sequence of the actions executed in the demonstration as the initial edges in the CTG as shown in~\figref{learning}. 
We adapt a seq2seq model from machine translation literature~\cite{luong2015effective} as our demo interpreter. We do not use a frame-based classifier, as we do not need accurate per-frame action classification. What is critical here is that the sequence of actions $A$ provides reasonable initial action order constraints (edges) to our conjugate task graph. 
We do assume the training demonstrations in $\mathbb{T}_{seen}$ come with the action sequence $A$ as supervision for our demo interpreter.
We only require this ``flat'' supervision for $\mathbb{T}_{seen}$, as opposed to the strong hierarchical supervision used in the previous work~\cite{xu2017neural}.

\vspace{1mm}
\noindent\textbf{Graph Completion Network (GCN).} Given a valid path (action sequence) from the demo interpreter, the goal is to complete the edges that are not observed in the demo. 
We formulate this as learning  graph state transitions~\cite{johnson2016learning,kipf2016semi}. Our GCN iterates between two steps: (i) edge update and (ii) propagation. 
Given the node embedding $\NE_{gcn}(n_i)$ for each node $n_i$, 
  the edge strengths are updated as:
\begin{equation}
\vspace{-1mm}
\small
    \mathcal{C}_{ij}^{t+1} = (1 - \mathcal{C}_{ij}^{t}) \cdot f_{set}(N^t_i, N^t_j) + \mathcal{C}_{ij}^{t} \cdot f_{reset}(N^t_i, N_j^t),
\end{equation}
where $\mathcal{C}_{ij}^{t}$ is the adjacency matrix of the previous iteration, $f_{set}$ and $f_{reset}$ are MLPs for setting and resetting the edge, and $N_i=\NE_{gcn}(n_i)$ is the node embedding for node $i$. Given $\mathcal{C}^t$ and the current node embeddings $N^t$, the propagation step updates the node embeddings with:  
\begin{equation}
\vspace{-1mm}
\small
    N^{t+1}_i = rnn(a_i, N^t_i), a_i = \sum_{j} \mathcal{C}^t_{ij} f_{f}(N^t_j) + \mathcal{C}^t_{ji} f_{b}(N^t_j), 
\end{equation}
where $rnn(a_i, N^t_i)$ takes the message $a_i$ from other nodes as input and updates the hidden state $N^t_i$ to $N^{t+1}_i$.

\subsection{Neural Task Graph Execution}
\label{sec:executor}

We have discussed how the NTG generates a CTG as the compositional representation of a task demonstration.
Next we show how to instantiate a policy from this task graph. We propose the NTG \textit{\exec} that interacts with the environment by executing the task graph.
The \exec executes a task graph in two steps:
(i) Node Localization: The \exec first localizes the current node in the graph based on the visual observation. 
(ii) Edge Classification: For a given node, there can be multiple outgoing edges for transitions to different actions. The  edge classifier checks the (latent) preconditions of each possible next action and picks the most fitting one.
These two steps enable the \exec to use the generated Conjugate Task Graph as a reactive policy which completes the task given observations. Formally, we decompose this policy as: 
$\pi(a|o) \propto \epsilon(a|n, o)\ell(n|o)$, 
where the localizer $\ell(n|o)$ localizes the current node $n$ based on visual observation $o$, and the edge classifier $\epsilon(a|n, o)$ classifies which edge transition from $n$ and $o$. Deciding the edge transition given the node is equivalent to selecting the next action $a$.

\vspace{1mm}
\noindent\textbf{Node Localizer.} % The first step of our execution is to localize the current node in the graph. 
We define the localizer as:
%\begin{equation}
    $\ell(n|o) \propto Enc(o)^T \NE_{loc}(n)$,
%\end{equation}
where the probability of a node is proportional to the inner product between $Enc(o)$, the encoded visual observation, and $\NE_{loc}(n)$, the node embedding of the node. 
Since our nodes are actions that are already observed in the seen tasks, we can learn the node embeddings effectively. This shows the benefit of modularizing our policy, where sub-modules are more generalizable.

\vspace{1mm}
\noindent\textbf{Edge Classifier.} The edge classifier is the key for NTG to generalize to unseen tasks. Unlike the localizer, which is approx. invariant across seen and unseen tasks, deciding the correct edge requires the edge classifier to correctly infer the underlying states from the visual observations. Take block stacking as an example. For a task that aims to stack blocks A, B, and C in order, the robot should not pick-and-place C unless B is already \textit{on A}. The edge classifier thus needs to recognize such prerequisites for actions involving block C. 
\begin{equation}
\vspace{-1mm}
\small
    \epsilon(a|n,o) \propto (W_\epsilon [Enc(o), \NE_{gcn}(n)])^T \NE_{loc}(n_a),
% \vspace{-1mm}
\end{equation}
where $n_a$ is the node for action $a$, and $\NE_{gcn}(\cdot)$ is the final node embedding from our GCN in \secref{generator}. As the GCN node embedding is used to generate edges in the conjugate task graph, it captures the task structure. We use $\NE_{loc}$ from localization for the destination node.

\subsection{Learning NTG Networks}
\label{sec:learning}

\begin{figure}[t!]
    \centering
     \includegraphics[width=\linewidth]{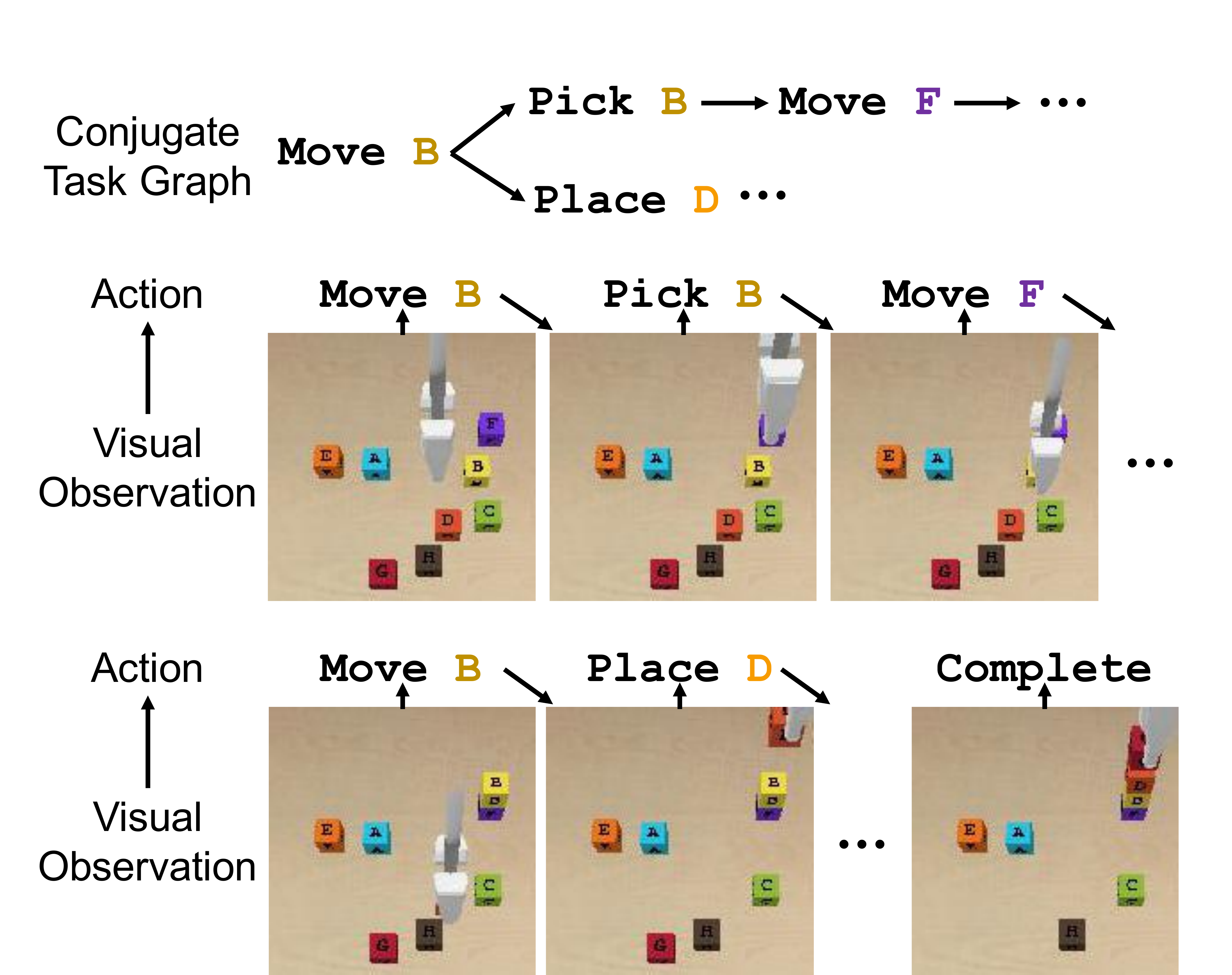}
    \caption{Execution of NTG based on the conjugate task graph. Although the \exec visited the \texttt{(Move B)} node twice, it is able to correctly decide the next action using the edge classifier by understanding the second visit needs to \texttt{(Place D)}.}
    \label{fig:ntp_qual}
\end{figure}

We have described how we decompose $\phi(\cdot)$ into the generator and the \exec. As discussed in \secref{problem} we train both on $\mathbb{T}_{seen}$.
In contrast to previous works that require strong supervision on $\mathbb{T}_{seen}$ (state-action pairs~\cite{duan2017one} or hierarchical supervision~\cite{xu2017neural}), NTG only requires the raw visual observation along with the flat action sequence (lowest level program in~\cite{xu2017neural} without a manually defined action hierarchy). An overview of learning different components of NTG is shown in \figref{learning}.

\vspace{1mm}
\noindent\textbf{Learning Graph Generation.} 
For each demonstration $d^{\tau}_i$ of task $\tau$, we have the corresponding $A^{\tau}_i = [a_1, \dots, a_K]$, the executed actions. First, we translate $A_i$ to a path $\{P^{\tau}_i = (\tilde{V}, \tilde{E}^{\tau}_i)\}$ by using all actions as nodes $\tilde{V}$ and adding edges of the transitions in $A_i$ to $\tilde{E}_i$. For a single task $\tau$, we use the union of all demonstrated paths of $\tau$ as the edges $E_t = \bigcup_{i} \tilde{E}^{\tau}_i$ of the groud truth conjugate task graph $g_{\tau} = (V, E_t)$. In this case, the goal of GCN is to transform each $P^{\tau}_i$ to $g_{\tau}$ by completing the missing edges in $P^{\tau}_i$. We use the binary cross entropy loss following~\cite{johnson2016learning} to train the GCN, where the input is $P^{\tau}_i$ and the goal is to generate  $g_{\tau}$.

\vspace{1mm}
\noindent\textbf{Learning Graph Execution.} Given a task graph from the generator, we learn an \exec that derives the policy. As discussed in~\secref{executor}, we decompose the policy into node localizer and edge classifier. 
For the localizer, we use the video frames as input and the corresponding action labels from the demonstrations as targets. For the edge classifier, we collect all pairs of source-target nodes connected by transitions, and use the action label from the demonstration as the target. Additionally, the edge classifier uses the  node embedding from our Graph Completion Network. The idea is that the embedding from the GCN can inform the edge classifier about what kind of visual state it should classify and learn to generalize to the unseen task.

%% file: 4_experiment.tex
\begin{figure}[!t]
    \centering
    \includegraphics[width=0.95\linewidth]{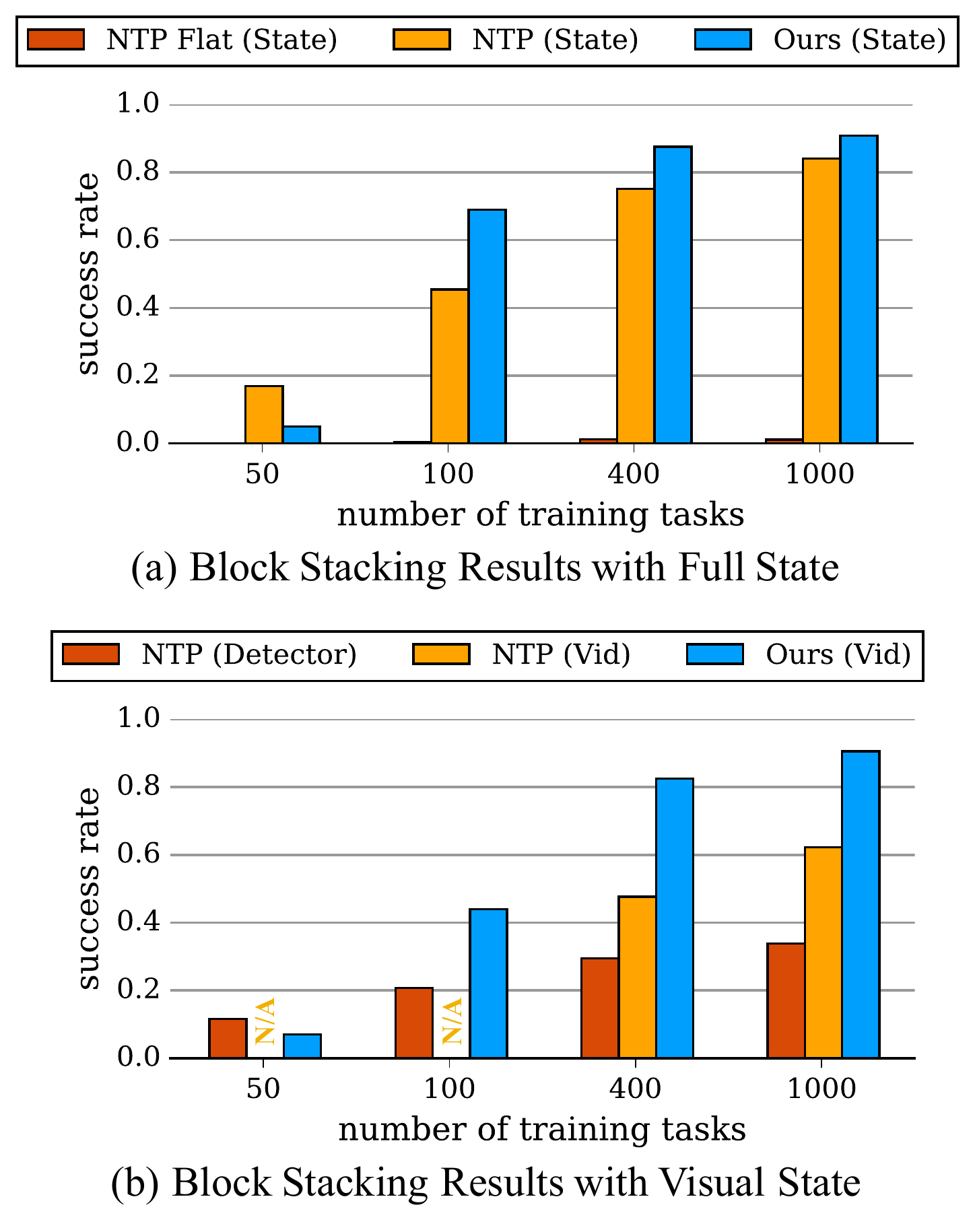}
    \caption{ Results for generalizing block stacking to unseen target configuration. (a) Results with the block locations as input, and (b) Results with raw video as input. Our NTG model significantly outperforms the baselines despite using only flat supervision. 
    }
    \label{fig:ntp_exp}
    \vspace{-8pt}
\end{figure}

\section{Experiments}
% \todo{DETAILS}

Our experiments aim to answer the following questions: 
(1) With a single \emph{video} demonstration, how does NTG generalize to unseen tasks and compare to baselines without using compositionality? (2) How do each of the components of NTG contributes to its performance? (3) Is NTG applicable to real-world data? For the first two questions, we evaluate and perform ablation study of NTG in two challenging task domains: the Block Stacking~\cite{xu2017neural}  using the BulletPhysics~\cite{BulletPhysics} and the Object Collection task in the AI2-THOR~\cite{zhu2017target}. For the last question, we evaluate NTG on real-world surgical data and examine its graph prediction and evaluation of unseen tasks on the JIGSAWS \cite{Gao2014JHUISIGA} dataset.

\subsection{Evaluating Block Stacking in BulletPhysics}
\label{sec:block}

We evaluate NTG's generalization to unseen target configurations. The hierarchical structure of block stacking provides a large number of unique tasks and is ideal for analyzing the effect of explicitly introducing compositionality.

\vspace{1mm}
\noindent\textbf{Experimental Setup.} 
The goal of Block Stacking is to stack the blocks into a target configuration. We follow the setup in Xu \etal~\cite{xu2017neural}. 
We use eight 5\,cm cubes with different colors and lettered IDs.
A task is considered successful if the end configuration matches the task demonstration. 
We use the 2000 distinct Block Stacking tasks and follow the training/testing split of Xu \etal~\cite{xu2017neural}.

\vspace{1mm}
\noindent\textbf{Baselines.} We compare to the following models:

\noindent\textit{ - Neural Task Programming (NTP)}~\cite{xu2017neural} learns to synthesize policy from demonstration by
decomposing a demonstration recursively. 
In contrast to ours, NTP assumes strong structural supervision: both the program hierarchy and the demonstration decomposition are required at training.
We use NTP as an example of methods that encourage compositionality via strong structural supervision.

\noindent\textit{ - NTP Flat} is an ablation of NTP, which only uses the same supervision as our NTG model (lowest level program).

\noindent\textit{ - NTP (Detector)}  first detects the block and feeds that into the model as the approximated full state. The detector is trained separately with additional supervision.

\begin{figure}[t]
    \centering
    \includegraphics[width=0.95\linewidth]{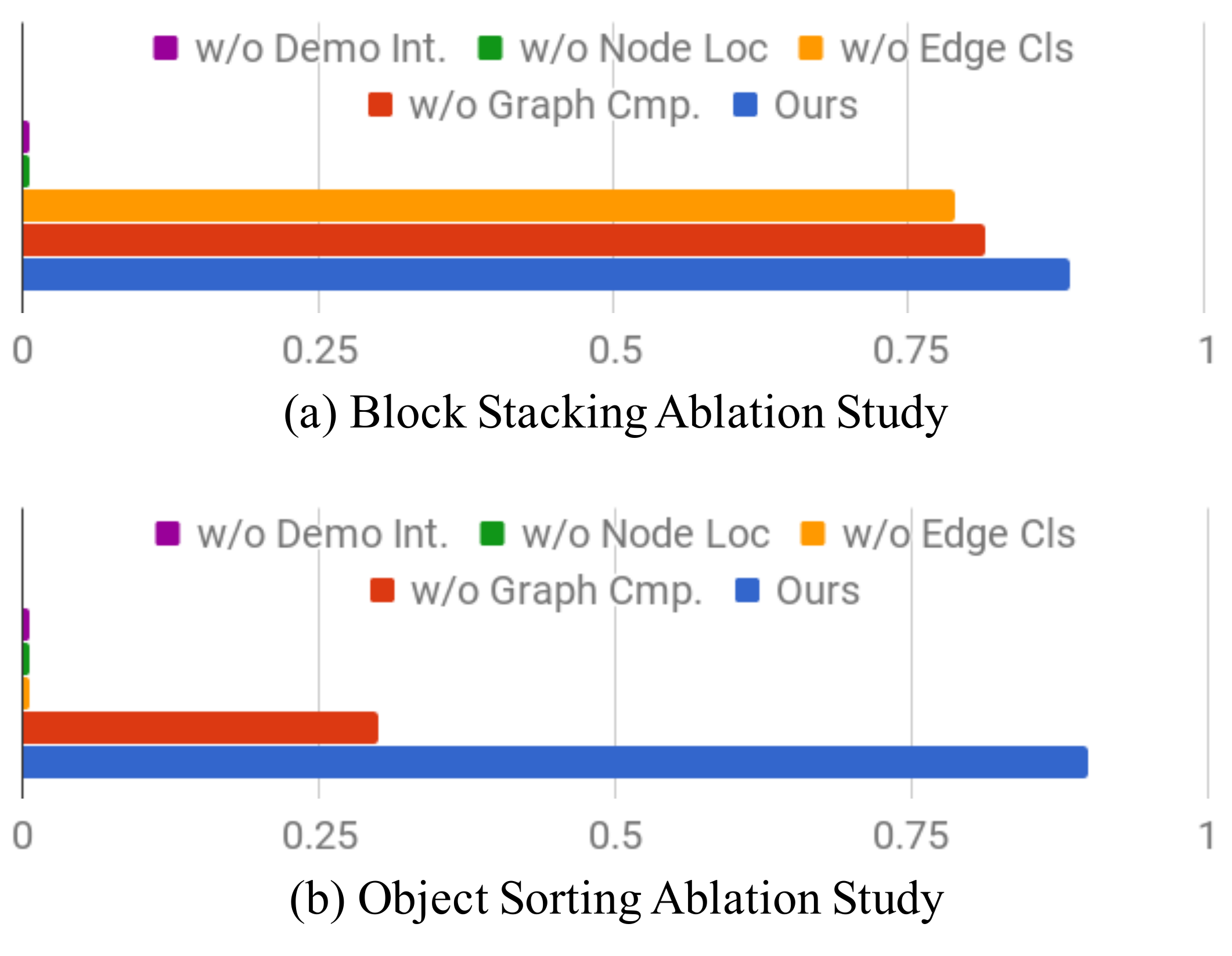}
    \caption{ Ablation study of NTG. (a) Demo Int. and Node Loc. are almost indispensable. (b) Both GCN and Edge Cls are required to generalize to  execution order different from the demonstration.
    }
    \label{fig:ablation}
\end{figure}

\vspace{1mm}
\noindent\textbf{Results.} Results are shown in \figref{ntp_exp}. The x-axis is the number of training seen tasks. We compare models with full state (State) and visual state (Vid) as input. Full state uses the 3D block location, 
and the visual state uses $64\times64$ RGB frames. 
For both input modalities, NTG can capture the structure of the tasks and generalize better to unseen target configuration compared to the baseline. NTG with raw visual input (Ours (Vid)) performs on-par with NTP using full state (NTP (State)). 
When there is not enough training data (50 tasks), the NTP (State) and NTP (Detector) in  are able to outperform NTG because of the extra supervision (hierarchical for NTP (State), and detection for NTP (Detector)).
However, once NTG is trained with more than 100 tasks, it is able to quickly interpret novel tasks and significantly outperforms the baselines. \figref{ntp_qual} shows an NTG execution trace. Although the \exec visited the \texttt{(Move B)} node twice, it is able to correctly decide the next action based on the visual observation by interpreting the underlying state from the visual observation.

\begin{figure}[t!]
    \centering
    \includegraphics[width=1\linewidth]{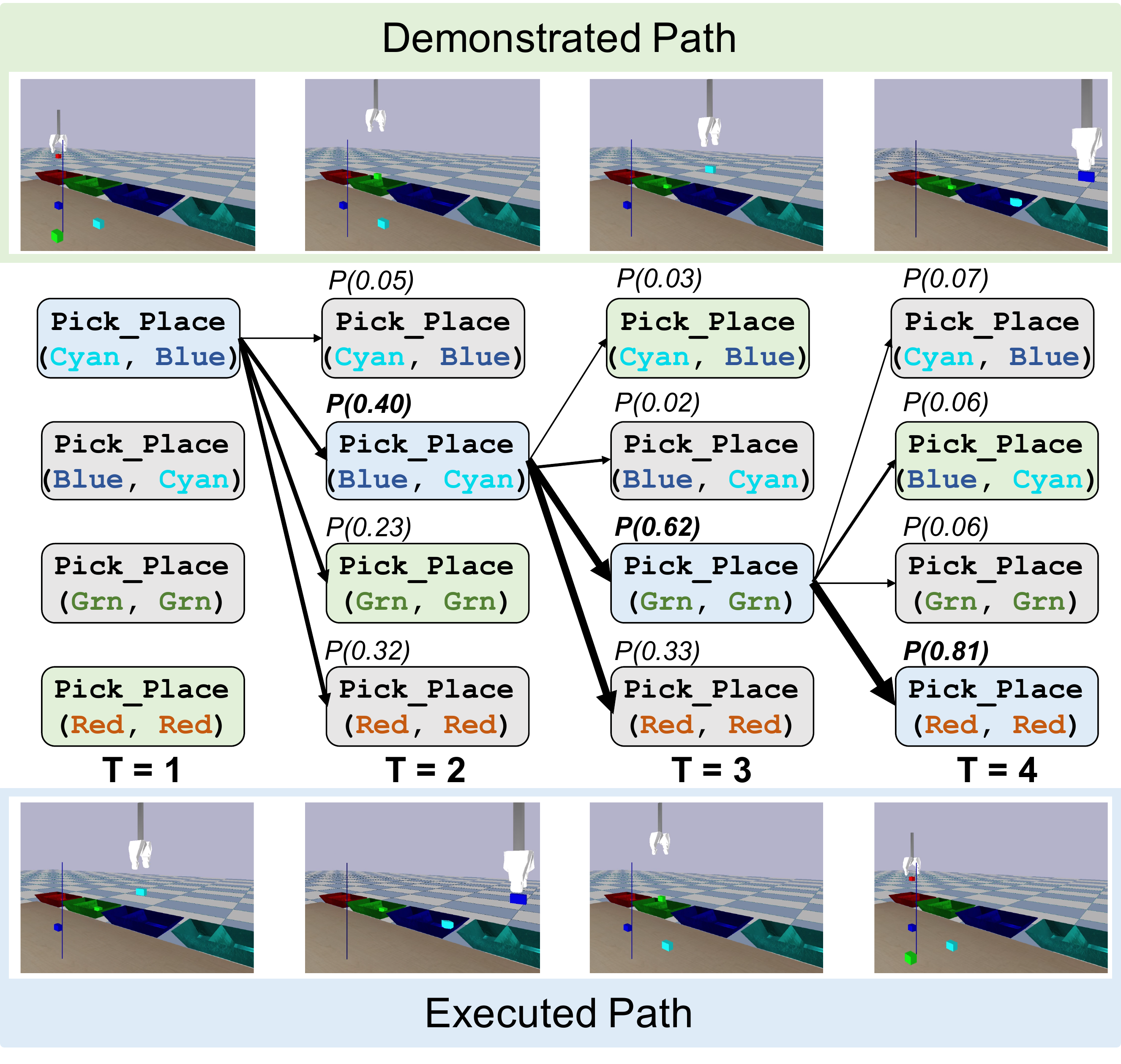}
    \caption{Using GCN, our policy is able to solve an unseen sorting task in a different order than the provided demonstration.
    }
    \label{fig:sortingntg}
\end{figure}

\begin{figure*}[t]
    \centering
    \includegraphics[width=1.0\linewidth]{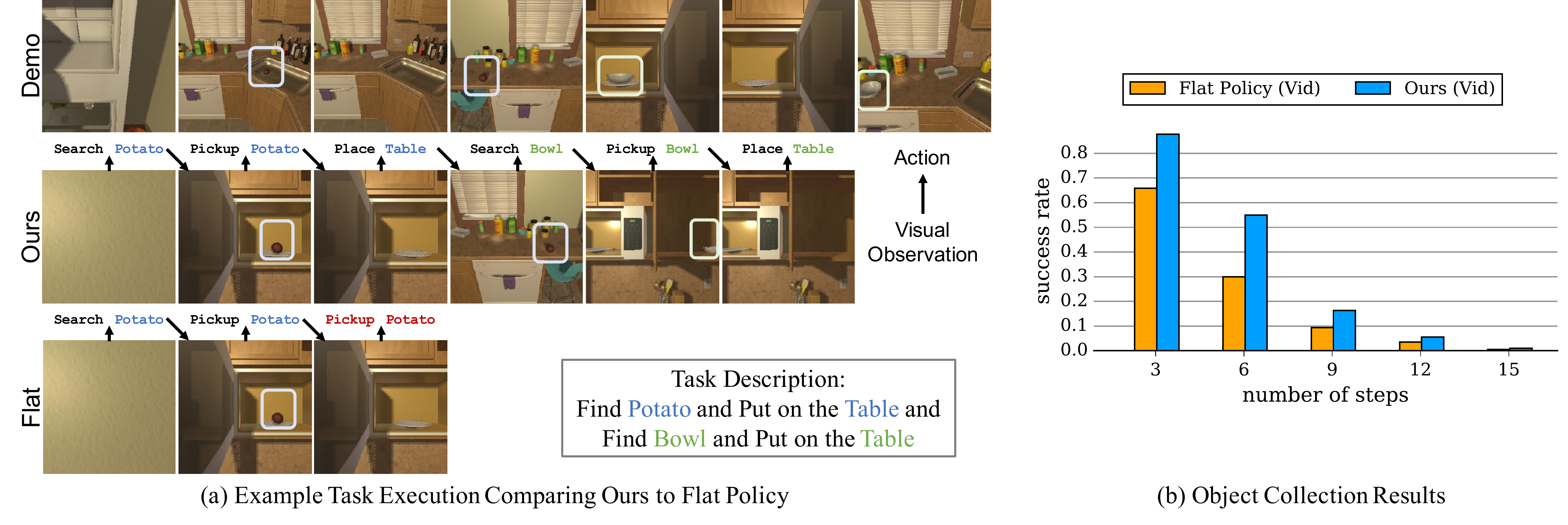}
    \caption{(a) Object Collection results. The bnding boxes are only for visualization and is not used anywhere in our model. The objects can appear in locations that are different from the demonstration, which leads to challenging and diverse visual state. NTG is able to understand the underlying state (\eg if the object is found) from the visual input and successfully complete the task. (b) Object Collection results on varying numbers of steps. The NTG model is only trained with 6 and 12 steps, and is able to generalize well to other numbers of steps. 
    }
    \label{fig:thor_all}
    \vspace{-2mm}
\end{figure*}
\subsection{Ablation Analysis of NTG Model Components}

Before evaluating other environments, we analyze the importance of each component of our model. Some sub-systems are almost \emph{indispensable}. For example, without the Demo Interpreter, there is no information from the video demonstration, and the policy is no longer task-conditional. 
We perform the ablation study using 1000 training tasks as follows: For Demo Interpreter, we initialize CTG as a fully connected graph without order constraints from the demonstration. For Node Localizer and Edge Classifier, we replace the corresponding term in the policy $\pi(a|o) \propto \epsilon(a|n, o)\ell(n|o)$ by a constant. For GCN, we skip the graph completion step. As shown in \figref{ablation}(a), the policy cannot complete any of the tasks without Demo Interpreter or Node Localizer. While our full model still performs the best, removing Edge Classifier or GCN does not give as big a performance gap. This is because the Block Stacking tasks from \cite{xu2017neural} do not all require task structure understanding.

\noindent\textbf{Alternate Solutions for Task.} GCN is particularly important for situations requiring alternative execution orders. For example, the task of ``putting the red ball into the red bin and the blue ball into the blue bin''. It is obvious to us that we can either put the red ball first or the blue ball first. This ability to generalize to alternative execution orders is exactly what we aim to capture with GCN. 
Without GCN, the policy can be easily stuck at unseen execution order (\emph{i.e., }not understanding object sorting order can be swapped). 
We thus analyze GCN on the ``Object Sorting'' task (details in Section VI of \cite{xu2017neural}), but initialize the scene to require execution order different from the demonstration. These settings will occur often when the policy needs to recover from failure or complete a partially completed tasks.  This is challenging because: (i) GCN has to generalize and introduce alternative execution order beyond the demonstration. (ii) Edge Classifier needs to correctly select the action from the newly introduced edges by GCN. As shown in \figref{ablation}(b), the policy cannot complete any of the tasks without Edge Classifier because of the ambiguities in the completed task graph.
\figref{sortingntg} shows an qualitative example of how our method learns to complete ``Object Sorting'' with order different from the demonstration using GCN. This shows the importance of both the Edge Classifier and GCN, which are required to complete this challenging task.

\iffalse
\begin{figure}[t]
    \centering
    \includegraphics[width=0.75\linewidth]{figs/thor_exp_plot.pdf}
    \caption{
    Object Collection results on varying numbers of objects. The NTG model is only trained with 2 and 4 objects, and is able to generalize well to other numbers of objects. 
    }
    \label{fig:thor}
\end{figure}
\fi

\begin{figure*}[t!]
    \centering
    \includegraphics[width=0.95\linewidth]{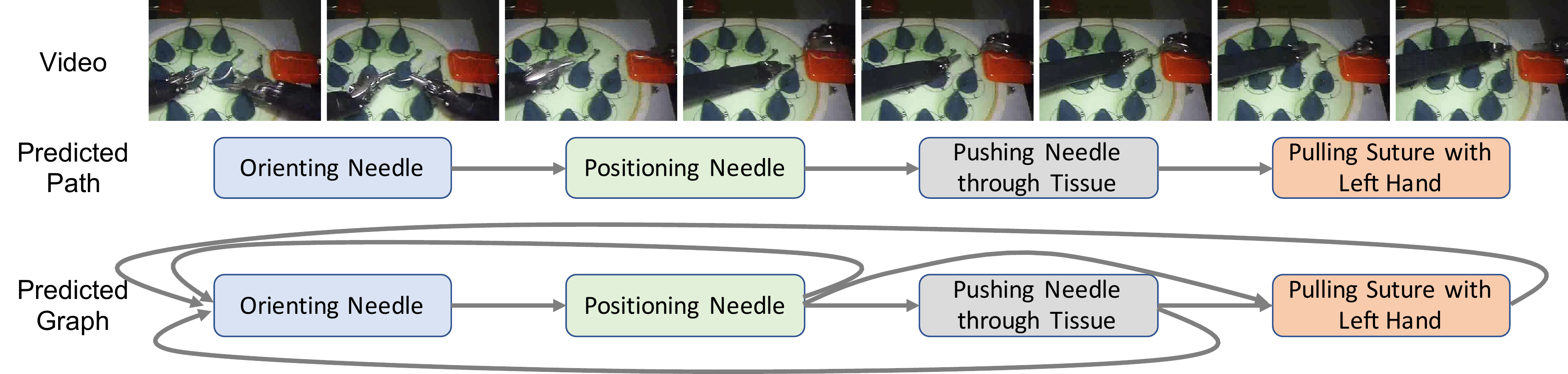}
    \caption{Part of a predicted graph from a single demonstration of an unseen task on the JIGSAWS dataset. Our method is able to learn that for the Needle Passing task, after failing any of the step in this subtask, the agent should restart by reorienting the needle.
    }
    \label{fig:jigsawgraph}
    \vspace{-1mm}
\end{figure*}

\subsection{Evaluating Object Collection in AI2-THOR}

In this experiment, we evaluate the Object Collection task, in which an agent collects and drop off objects from a wide range of locations with varying visual appearances. We use AI2-THOR~\cite{zhu2017target} as the environment, which allows the agent to navigate and interact with objects via semantic actions (\eg \texttt{Open}). %, and the state changes are rendered by graphics engine accordingly.
This task is more complicated than block stacking be: First, the agent is navigating in the scene and thus can only have partial observations. Second, the photo-realistic simulation enables a variety of visual appearance composition. In order to complete the task, the model needs to understand various appearances of the object and location combinations.

\vspace{1mm}
\noindent\textbf{Experimental Setup.} An Object Collection task involves visiting $M$ randomly selected searching locations for a set of $N$ target objects out of $C$ categories. Upon picking up a target object, the agent visits and drops off the object at one of $K$ designated drop-off receptacles. A task is considered successful if all of the target objects are placed at their designated receptacles at the end of the task episode. The available semantic actions are \texttt{search}, \texttt{pickup(object)}, \texttt{dropoff(receptacle)}. The \texttt{search} action visits each searching locations in a randomized order.  \texttt{pickup(object)}  picks up a selected object and the action would fail if the selected object is not visible to the agent. \texttt{dropoff(receptacle)} would teleport the agent to a selected drop-off receptacle (\texttt{tabletop, cabinet}, etc) and drop off.  We use $N=[1, 5]$ objects (3-15 steps) out of $C=8$ categories, $M=N+3$ search locations, and $K=5$ drop-off receptacles. % One example demonstration is shown in \figref{thor}(a).

\vspace{1mm}
\noindent\textbf{Baseline.} We compare to the ``Flat Policy'' baseline in~\cite{duan2017one} to show the importance of incorporating compositionality to the policy. 
At each step, the Flat Policy uses attention to extract relevant information from the demonstration and combine it with the observation to decide action.
For a fair comparison, we implement the Flat Policy using the same architecture as our demo interpreter.
Note that the Object Collection domain doesn't have hand-designed hierarchy. Hence NTP~\cite{xu2017neural} is reduced to a similar flat policy model. % without the hierarchical supervision.

\vspace{1mm}
\noindent\textbf{Results.} The results for Object Collection on different numbers of objects are shown in \figref{thor_all}(b). The models are only trained on 2 and 4 objects and generalize to 1, 3, 5 objects.
NTG significantly outperforms the Flat Policy on all numbers of objects. This shows the importance of explicitly incorporating compositionality. Qualitative comparison is shown in \figref{thor_all}(a). The bounding boxes are for visualization only and are not used in the model. 
During evaluation, the objects of interest can appear in locations that are different from the demonstration and thus lead to diverse and challenging visual appearances.
It is thus important to understand the structure of the demonstration instead of naive appearance matching. Our explicit model of the task structure sets NTG apart from the flat policy and leads to stronger generalization to unseen tasks.

\subsection{Evaluating Real-world Surgical Data}

\begin{figure}[t]
    \centering
    \includegraphics[width=0.95\linewidth]{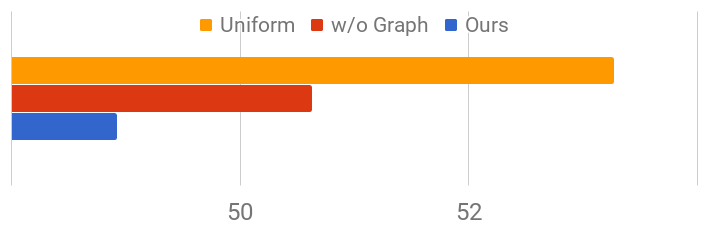}
    \caption{Negative loglikehood (NLL) of expert demonstrations on the JIGSAWS dataset. The policy generated by our full model can best capture the actions performed in human demonstration. 
    }
    \label{fig:jigsaws}
    \vspace{-1mm}
\end{figure}

We have shown that NTG significantly improves one-shot visual imitation learning by explicitly incorporating compositionality. We now evaluate if this structural approach can be extended to the challenging real-world surgical data from the JIGSAWS dataset \cite{Gao2014JHUISIGA}, which contains videos and states for surgical tasks, and the associated atomic action labeling. In this setting, our goal is to assess NTG's ability to generalize to the task of ``Needle Passing'', after training on the tasks of ``Knot Tying'' and ``Suturing''. This is especially challenging because it requires generalization to a new task with significant structural and visual differences, given only 2 task types for training.

Without a surgical environment, we cannot directly evaluate the policy learned by NTG on the JIGSAWS dataset. Therefore, we evaluate how well the NTG policy is able to predict what a human will do in other demonstrations. This entails generating a policy conditioned on a single demonstration of ``Needle passing'', and using it to evaluate the negative log-likelihood (NLL) of all the other demonstrations in the ``Needle Passing'' task. A lower negative log likelihood corresponds to the generated policy better explaining the other demonstrations, and in turn better capturing the task structure. 

The results are shown in \figref{jigsaws}. We compare to the no graph variant of our model and also the lower bound of a uniform policy. Unsurprisingly, the \textit{uniform} policy performs the worst without capturing anything from the demonstration. The \textit{no-graph} variant is able to capture some parts of the expert policy and better capture the expert demonstration. However, the policy generated by full NTG model substantially improves the NLL and is the most consistent with the expert demonstration. 

In addition, we show qualitative results of part of our task graph prediction on the JIGSAWS dataset in \figref{jigsawgraph}. Again, we train on ``Knot Tying'' and ``Suturing'' and evaluate on ``Needle Passing''. By comparing the predicted path and the final predicted graph, we can see that our model is able to introduce several new edges going back to the action ``Orienting Needle''. This captures the behavior that when the execution fails in any of step in this subtask of ``Needle Passing'', the agent should return to  ``Orienting Needle'' and reorient the needle to restart the subtask. This is consistent with our intuition and the ground truth graph.

%% file: 5_conclusion.tex
\section{Conclusion}

We presented Neural Task Graph (NTG) Network, a one-shot visual imitation learning method that explicitly incorporates task compositionality into both the intermediate task representation and the policy. Our novel Conjugate Task Graph (CTG) generation and execution formulation effectively handles unseen visual states and endows the method with a reactive and executable policy. We demonstrate that NTG is able to outperform both methods with unstructured representation~\cite{duan2017one}, and methods with a hand-designed hierarchical structure~\cite{xu2017neural} on a diverse set of tasks, including simulated environment with photo-realistic rendering and a real-world dataset. 